\documentclass[]{metaai/fairmeta}

\usepackage{microtype}
\usepackage{graphicx}
\usepackage{subfigure}
\usepackage{booktabs} 

\usepackage{hyperref}

\usepackage{amsmath}
\usepackage{amssymb}
\usepackage{mathtools}
\usepackage{amsthm}
\usepackage{enumitem}
\usepackage{wrapfig}

\usepackage{algorithmic}

\usepackage{varwidth}
\usepackage{tikz}
\usetikzlibrary{
    positioning,
    arrows,
    arrows.meta
}
\tikzset{>={Latex[width=2mm,length=2mm]}}
\usepackage{multirow}

\def\ours{Searchformer}
\def\planonly{solution-only}
\def\Planonly{Solution-only}
\def\plantrace{search-augmented}
\def\Plantrace{Search-augmented}

\def\PlanonlyNoHyphen{Solution only}

\def\PlantraceNoHyphen{Search augmented}
\def\planimproved{Searchformer}

\title{Beyond \texorpdfstring{$A^*$}{A*}: Better Planning with Transformers via Search Dynamics Bootstrapping}

\author[1]{Lucas Lehnert}
\author[1]{Sainbayar Sukhbaatar}
\author[1]{DiJia Su}
\author[1]{Qinqing Zheng}
\author[1]{Paul Mcvay}
\author[1]{Michael Rabbat}
\author[1]{Yuandong Tian}

\affiliation[1]{FAIR at Meta}

\abstract{
    While Transformers have enabled tremendous progress in various application settings, such architectures still trail behind traditional symbolic planners for solving complex decision making tasks.
    In this work, we demonstrate how to train Transformers to solve complex planning tasks.
    This is accomplished by training an encoder-decoder Transformer model to predict the \emph{search dynamics} of the $A^*$ search algorithm.
    We fine tune this model to obtain a \emph{\textbf{\ours{}}}, a Transformer model that optimally solves previously unseen Sokoban puzzles 93.7\% of the time, while using up to 26.8\% fewer search steps than the $A^*$ implementation that was used for training initially. 
    In our training method, $A^*$'s search dynamics are expressed as a token sequence outlining when task states are added and removed into the search tree during symbolic planning. 
    \ours{} significantly outperforms baselines that predict the optimal plan directly with a 5--10$\times$ smaller model size and a 10$\times$ smaller training dataset. 
    Lastly, we demonstrate how \ours{} scales to larger and more complex decision making tasks with improved percentage of solved tasks and shortened search dynamics.
}

\correspondence{\email{\{lucaslehnert, yuandong\}@meta.com}}

\metadata[Code]{\url{https://github.com/facebookresearch/searchformer}}

\begin{document}

\maketitle

\section{Introduction}
\label{sec:introduction}

Transformer-based architectures~\citep{vaswani2017transformer} have demonstrated impressive performance in different tasks, including holding conversations at the human level~\citep{Shuster2022BlenderBot3A,openai2022chatgpt,openai2023gpt4,touvron2023llama2}, high-quality image understanding~\citep{caron2021dino,oquab2024dinov2,assran2023ijepa} and video generation~\citep{singer2023makeavideo}, multi-modal generation~\citep{girdhar2023emu,radford2021openclip}, and code completion~\citep{roziere2023codellama,openai2021codex}. 
By training these architectures on internet-scale datasets, the resulting models, such as Large Language Models (LLMs), can generalize well in real-world use cases. 

Despite these successes, Transformer-based architectures and LLMs still struggle when it comes to solving planning and reasoning tasks. 
Previous studies demonstrate that LLMs fall short in multi-step planning tasks~\citep{valmeekam2023planningstudy,valmeekam2023llmplanning} or when performing higher-order reasoning~\citep{momennejad2023evaluatingllms,fan2020tranformersfeedbackmemory}.

In recent years, various methods have been proposed to improve the performance of Transformers in these settings. 
One approach is to simulate the human thinking process and produce intermediate ``thoughts'' before outputting a response. 
Chain-of-Thought (CoT) prompting~\citep{wei2022chainofthought} and the Tree-of-thoughts (ToT) method~\citep{yao2023tot} encourage the model to ``think'' step by step. 
While these techniques are often effective, they can also lead to worse performance, for example due to self-enforcing~\citep{huang2023llmselfcorrectreasoning}. 
Furthermore, techniques effective on one dataset may not work well on others due to changes in the type of reasoning involved (e.g., spatial reasoning vs. mathematical reasoning). 
How to enable Transformers and LLMs to plan, solve multi-step decision making tasks, and perform reasoning still remains elusive and an active area of research.

\subsubsection*{Our work}

We demonstrate how to train Transformers to robustly solve complex planning tasks.
Similar to LLMs, we train Transformers to predict the next word given a sequence of words.
Our experiments use synthetically generated datasets with a synthetic language and vocabulary.
Using this framework, we demonstrate how to construct training data such that the resulting model imitates the computation performed by $A^*$ search~\citep[Chapter 3]{russelnorvig2021ai}.
Lastly, we present \emph{\ours{}}, a Transformer model that solves complex planning tasks in fewer search steps than our $A^*$ reference implementation.
This model is obtained through \emph{search dynamics bootstrapping}, a method that first trains a Transformer to imitate $A^*$'s search process and then fine-tunes the model to find an optimal plan within fewer search steps.

To train a Transformer to perform planning, we express a planning task and its optimal solution plan as a sequence of words, called \emph{tokens}.
We also log the computation performed by $A^*$ into an execution trace consisting of a token sequence, resulting in a sequence dataset that captures $A^*$'s \emph{search dynamics}.
Using these \emph{\plantrace{}} sequences, a Transformer model is trained to generate token sequences that encode $A^*$'s search dynamics along with an optimal plan. 

Subsequently, the resulting \plantrace{} model is fine-tuned to generate shorter token sequences while still outputting an optimal plan.
We refer to this final fine-tuned model as a \ours{}.
When solving Sokoban puzzles, our models solve 93.7\% of all test tasks while performing on average 26.8\% fewer search steps than our $A^*$ reference implementation.

Through a sequence of experiments that control task complexity, dataset size, and model size, we demonstrate that including execution traces into the training data increases performance on an independent test task set---despite a $10\times$ to $100\times$ increase in the length of the generated sequences.
We find that \plantrace{} models (that include execution traces into their training data) generate an optimal plan more often on unseen tasks with ten times fewer training sequences than a larger \planonly{} model (that is trained on sequences only including a task description and task solution).
This result highlights the power of including $A^*$'s search dynamics into the training process of Transformer models. 

\section{Related Work}

While existing work~\citep{trinh2024alphageometry,ruoss2024grandmasterlevel} leverages synthetic datasets to learn policies for reasoning, our study is fundamentally different in this regard. 
We focus on improving the reasoning capability embedded in a Transformer's weights. 
Existing algorithms such as AlphaZero~\citep{silver2018alphazero}, MuZero~\citep{schrittwieser2020muzero}, and AlphaGeometry~\citep{trinh2024alphageometry} optimize a neural network using the output of existing symbolic planning algorithms, whose internal states are not used (i.e., treated as black-boxes).
For example,~\citet{silver2017alphago} use MCTS as a policy improvement operator to update the neural network's weights.
In contrast, the presented search dynamics bootstrapping method uses a Transformer model to generalize towards more efficient search patterns and improves the model itself.
A planning algorithm (together with its internal search dynamics) is used to initially train a Transformer model. 

Prior work focuses on training a neural network on execution traces of reasoning tasks~\citep{nye2021scratchpad} or on training a neural network to predict an optimal action~\citep{yang2022procedurecloning,pallagani2022plansformer,ruoss2024grandmasterlevel}.
In contrast, we focus on training a Transformer to generate the entire search process of $A^*$ when computing an optimal plan.
Instead of only predicting a single action, our model predicts an entire multi-step plan to solve a task.

Our work bears some similarity with neuro-symbolic systems~\citep{graves2014neuralturingmachines,cai2017neuralprogrammingarchitecture}, which build differentiable architectures to mimic the functionality of existing symbolic systems. 
However, these methods use dedicated components (e.g., explicit memory components, built-in recursion), while \ours{} focuses on next-token prediction.
Here, \ours{} relies on generating long contexts and position embeddings~\citep{chen2023extending,peng2023yarn} to predict in optimal plan.
Ultimately, our work sheds light on how to build more general architectures that automatically learn a planning mechanism. 

Using Transformer architectures to solve complex sequential decision making tasks has been studied in prior work in a reinforcement learning (RL) setting~\citep{chen2021decisiontransformer,janner2021trajectorytransformer,laskin2023algorithmdistillation}.
However, this prior work presents different methods to modelling trajectories of trial and error interactions and focuses on predicting a next action, state, or rewards or a combination of them.
In contrast, we demonstrate how to use a Transformer to model the search steps involved in computing an optimal multi-step plan.
MCTSNet~\citep{guez2018learning} also attempts to learn the search procedure itself, but still hard-codes the MCTS search procedure~\citep{coulom2006mcts} into a neural network, which leads to quadratic backpropagation overhead and can only deal with up to 500 step rollouts, while our approach can deal with much longer search execution trace. 
We demonstrate that Transformers can not only imitate a symbolic planning algorithm but can also be used to discover more efficient heuristics via fine-tuning.

\begin{figure}
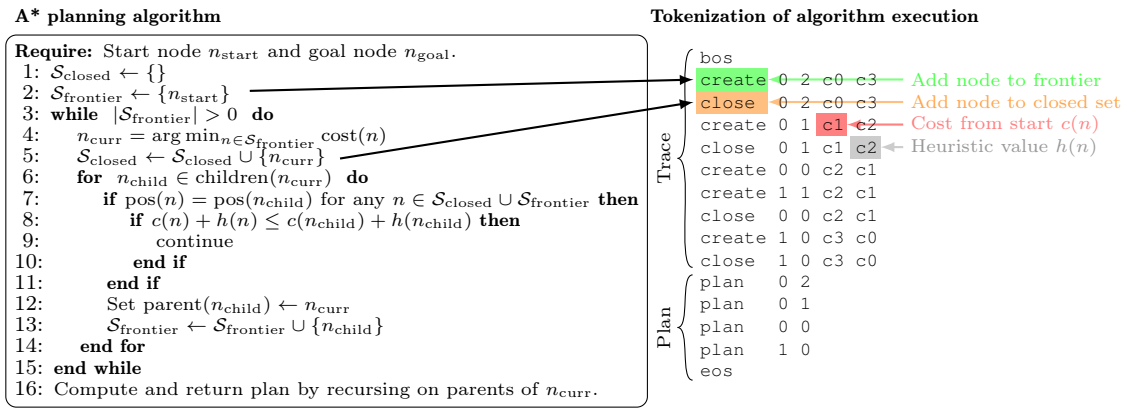

    \vspace{-2mm}
    \centering
    \subfigure[
        Maze navigation task
    ]{
        \label{fig:maze-example}
        \begin{tikzpicture}
            \fill[white, draw=lightgray,thick] (0.0,0.0) rectangle (0.5,0.5);
\fill[orange!50!white, draw=lightgray,thick] (0.5,0.0) rectangle (1.0,0.5);
\fill[lightgray, draw=lightgray,thick] (1.0,0.0) rectangle (1.5,0.5);

\fill[white, draw=lightgray,thick] (0.0,0.5) rectangle (0.5,1.0);
\fill[white, draw=lightgray,thick] (0.5,0.5) rectangle (1.0,1.0);
\fill[white, draw=lightgray,thick] (1.0,0.5) rectangle (1.5,1.0);

\fill[green!50!white, draw=lightgray,thick] (0.0,1.0) rectangle (0.5,1.5);
\fill[lightgray, draw=lightgray,thick] (0.5,1.0) rectangle (1.0,1.5);
\fill[white, draw=lightgray,thick] (1.0,1.0) rectangle (1.5,1.5);

\fill[blue!50!white] (0.25,1.25) circle (0.1);
\fill[blue!50!white] (0.25,0.75) circle (0.1);
\fill[blue!50!white] (0.25,0.25) circle (0.1);
\fill[blue!50!white] (0.75,0.25) circle (0.1);
\draw[-latex,draw=blue!50!white,thick] (0.25,1.15) -- (0.25,0.85);
\draw[-latex,draw=blue!50!white,thick] (0.25,0.65) -- (0.25,0.35);
\draw[-latex,draw=blue!50!white,thick] (0.35,0.25) -- (0.65,0.25);

\fill[lightgray, draw=lightgray,thick] (2.2,1.2) rectangle (2.5,1.5); \node[align=left,anchor=west] at (2.6, 1.35) {\small{: wall cell}};
\fill[green!50!white, draw=lightgray,thick] (2.2,0.7) rectangle (2.5,1.0); \node[align=left,anchor=west] at (2.6, 0.85) {\small{: start cell}};
\fill[orange!50!white, draw=lightgray,thick] (2.2,0.2) rectangle (2.5,0.5); \node[align=left,anchor=west] at (2.6, 0.35) {\small{: goal cell}};
\fill[blue!50!white] (2.25,-0.15) circle (0.1); \draw[-latex,draw=blue!50!white,thick] (2.35,-0.15) -- (2.65,-0.15); \node[align=left,anchor=west] at (2.6, -0.15) {\small{: plan step}};

\node[anchor=east] at (0,1.25) {\small{2}};
\node[anchor=east] at (0,0.75) {\small{1}};
\node[anchor=east] at (0,0.25) {\small{0}};

\node[anchor=north] at (1.25,0) {\small{2}};
\node[anchor=north] at (0.75,0) {\small{1}};
\node[anchor=north] at (0.25,0) {\small{0}};
        \end{tikzpicture}
    }~~~
    \subfigure[
        Tokenization of a planning task and its solution
    ]{
        \label{fig:planning-tokenization}
        \hspace{1.2cm}\renewcommand{\ttdefault}{pcr}
\renewcommand{\sfdefault}{cmr}
\begin{tikzpicture}
	\node[align=left,anchor=south west,font={\scriptsize}] at (5.2, 0) {\textbf{Prompt}};
	\node[font={\scriptsize}, anchor=north west] at (5.15,0.2) {
		\begin{tikzpicture}
			\input{figure/grid-prompt.tex}
		\end{tikzpicture}
	};
	\node[align=left,anchor=south west,font={\scriptsize}] at (7.8, 0) {\textbf{Response}};
	\node[font={\scriptsize}, anchor=north west] at (7.75,0.2) {
		\begin{tikzpicture}
			\input{figure/grid-plan.tex}
		\end{tikzpicture}
	};
\end{tikzpicture}\hspace{1.2cm}
    }
    \subfigure[
        $A^*$'s execution when solving a planning task is logged into an execution trace
    ]{
        \label{fig:astar-trace}
        \renewcommand{\ttdefault}{pcr}
\renewcommand{\sfdefault}{cmr}
\begin{tikzpicture}
	\node[align=left,anchor=south west,font={\scriptsize}] at (-1.25, 0) {\textbf{A* planning algorithm}};	
	\node[draw, rounded corners,font={\scriptsize}, anchor=north west] at (-1.25, 0) {
		\begin{varwidth}{0.5\linewidth}
			\input{figure/astar-alg.tex}
		\end{varwidth}
	};
	
	\node[align=left,anchor=south west,font={\scriptsize}] at (7.2, 0) {\textbf{Tokenization of algorithm execution}};
	\node[font={\scriptsize}, anchor=north west] at (7.2, 0) {
		\begin{varwidth}{0.5\linewidth}
			\begin{tikzpicture}
				\input{figure/astar-reasoning.tex}
			\end{tikzpicture}
		\end{varwidth}
	};
	
	\draw[-latex,thick] (2.0,-0.75) -- (7.9,-0.6);
	\draw[-latex,thick] (3.2,-1.65) -- (7.9,-0.9);
\end{tikzpicture}
    }
    \vspace{-4mm}
    \caption{
        \textbf{Expressing a planning task in token sequences.}
        \subref{fig:maze-example}: 
        A $3 \times 3$ maze navigation task where the goal is to find a the shortest path from start to goal without entering a wall cell.
        \subref{fig:planning-tokenization}: 
        The $3 \times 3$ maze navigation task is expressed as a prompt token sequence (left panel) and the optimal plan is expressed as a response token sequence (right panel).
        The start and end of a sequence is indicated by a beginning-of-sequence token, \texttt{bos}, and an end-of-sequence token, \texttt{eos}. 
        Numbers indicate $x,y$ coordinates.
        \subref{fig:astar-trace}:
        The search dynamics of the $A^*$ algorithm (left panel) is logged into an execution trace (right panel).
        The last two tokens in the trace encode the cost-since-start value $c(n)$ and the heuristic value $h(n)$ (letter ``c'' distinguishes costs from coordinate numbers).
        The $A^*$ algorithm is described in detail by \citet[Chapter 3]{russelnorvig2021ai}.
    }
    \label{fig:teaser}
\end{figure}

\section{Problem Setup}
\label{sec:planning-tokenization}

\autoref{fig:teaser} provides an overview of our synthetic dataset generation process.
We consider two domains: maze navigation (\autoref{fig:maze-example}) and solving Sokoban puzzles (\autoref{fig:sokoban-level} in Appendix~\ref{app:dataset-generation}). 
In maze navigation, the goal is to find the shortest path through an $n$-by-$n$ maze.
In Sokoban, a worker can move up, down, left, or right and has to push each box onto a dock to solve the puzzle.
An incorrect move may immediately lead to a dead end and thus reasoning across multiple time steps is required to solve the puzzle. 
Each state in a puzzle consists of a combination of box and worker positions, making Sokoban computationally more difficult to solve than maze navigation.

\subsection{Generating execution traces of \texorpdfstring{$A^*$}{A*} search.}

The $A^{*}$ algorithm computes an optimal plan by manipulating two sets of nodes:
\begin{itemize}[noitemsep,topsep=0pt,leftmargin=16pt]
    \item A frontier set containing the current search frontiers.
    \item A closed set containing all searched nodes. 
\end{itemize}
In the maze example in~\autoref{fig:maze-example}, each node corresponds to an empty (non-wall) grid cell.
For each node, the algorithm computes a heuristic value and a cost from start value.
At any given iteration, which node is searched next is determined by the content of the frontier and closed sets as well as the heuristic and cost from start values (\autoref{fig:astar-trace}, left panel).
$A^*$'s execution trace is collected by tracking all insertion operations into the frontier and closed set along with heuristic and cost from start values (\autoref{fig:astar-trace}, right panel).
The right panel in~\autoref{fig:astar-trace} illustrates the resulting trace for the maze example shown in~\autoref{fig:planning-tokenization}.
Each row corresponds either to an insertion of a node into the frontier, indicated by a \texttt{create} token, or to moving a node into the closed set, indicated by a \texttt{close} token.
Each node is represented by its $(x,y)$ position in the maze as well as the two cost tokens.
The resulting plan is then appended to this trace. 
This trace is constructed such that given any prefix the next token can be predicted correctly.
For the maze datasets, $A^*$ uses the Manhattan distance to the goal location as a heuristic.
In Sokoban, $A^*$ first matches every box to the closest dock and then computes the sum of all Manhattan distances between each box and dock pair.

For each experiment, we generate two token sequence variants, as illustrated in~\autoref{fig:teaser}: 
\begin{itemize}[noitemsep,topsep=0pt,leftmargin=16pt]
    \item \emph{\Planonly{} sequences} of the format \texttt{<prompt><plan>}, where the \texttt{<prompt>} part encodes a task description and the \texttt{<plan>} part the optimal plan (\autoref{fig:planning-tokenization}).
    \item \emph{\Plantrace{} sequences} of the format \texttt{<prompt><trace><plan>}, where the \texttt{<trace>} part encodes $A^*$'s execution trace (\autoref{fig:astar-trace}).
\end{itemize}

Because every model is trained from scratch, the resulting models are specifically trained to only predict sequences that outline optimal plans for a set of different planning tasks.
After training, the model's output is parsed and evaluated if it contains an optimal or feasible solution plan.

\vspace{-0.06in}
\subsection{Training a Transformer model}
\vspace{-0.06in}
When generating a token sequence dataset, each task is unique and the test set is constructed such that it does not contain any duplicate of the training set.
With this experiment design, we hope to gain insight into how Transformers can be used to solve planning tasks and generalize to previously unseen test tasks.

By including intermediate computation steps, the Transformer model is trained to effectively imitate the computation performed by the $A^*$ algorithm.
Different from Procedure Cloning~\citep{yang2022procedurecloning} where a neural network is learned to predict the optimal state/action sequence (in our case task prompts and optimal plans), our Transformer model also learns to predict the \emph{entire thinking process}, including the attempted but failed paths, that leads to the optimal plan. 

For each experiment an adaptation of the encoder-decoder T5 architecture~\citep{raffel2020t5} is used that integrates Rotary Position Embeddings (RoPE)~\citep{su2023rope}.
More details and hyper-parameters can be found in Appendix~\ref{app:hyper-parameters}.
The encoder processes the \texttt{<prompt>} part of a training sequence, and the decoder processes either a \texttt{<trace><plan>}-formatted sequence (\plantrace{} model) or only a \texttt{<plan>}-formatted sequence (\planonly{} model).
Depending on the model variant, each network is trained to maximize the cross-entropy between the distribution of the decoder generations and the distribution of sampling a corresponding sequence from the training dataset.
Appendix~\ref{app:methods} describes our optimization setup in more detail.

\subsection{Moving past algorithm imitation via search dynamics bootstrapping}
\label{sec:bootstrapping}

To reduce the number of tokens generated by a \plantrace{} model during inference, we implement a method to shift the distribution with which the decoder generates execution traces.
First, a \plantrace{} model is trained to imitate the search dynamics of $A^*$ search.
To discover new search dynamics with this \plantrace{} model and explore the execution trace space, the \plantrace{} model must generate different sequences for the same task prompt.
We accomplish this by inducing non-determinism into the training data and use a non-determinsitic $A^*$ implementation that breaks cost ties randomly and randomizes the order with which child nodes are expanded.
This approach does not decrease the efficiency of $A^*$ search itself and merely changes the order with which different nodes are searched while still respecting $A^*$'s heuristic and cost calculations.
The resulting \plantrace{} model will then approximate the probability distribution with which the training sequences were generated.

Once a model is trained to imitate the search dynamics of non-deterministic $A^*$ search, it is used to generate a \emph{new} training dataset consisting of shorter token sequences.
This new dataset is constructed by using the trained \plantrace{} model to sample multiple different token sequences for each training prompt.
In this step, we only use the training dataset for bootstrapping and not the test dataset.
Each generated sequence is parsed and checked if it ends in an optimal plan.
If this is the case and the sequence is also shorter than the corresponding sequence contained in the original training dataset, then this shortened sequence is included in the new short sequence training dataset.
If the generated sequence does not end in an optimal plan or is longer than the original training sequence, then the sequence from the original training dataset is re-used.

Subsequently, the \plantrace{} model is fine-tuned on the new short sequence training dataset.
To distinguish from the \plantrace{} model that imitates $A^*$'s search dynamics, we call this new model \emph{\planimproved{}}.
This procedure can then be repeated by using the resulting fine-tuned model to generate the next even shorter sequence dataset and then fine-tuning the \planimproved{} model again.
In Section~\ref{sec:trace-improvement} we demonstrate that this procedure does in fact reduce the number of steps performed during inference while further improving performance.
The \planimproved{} model no longer imitates $A^*$ search and has instead discovered a new way of solving a planning problem using fewer steps.

\section{Experiments}
\label{sec:experiments}

In our experiments, we use two different $A^*$ implementations for sequence data generation:

\begin{enumerate}[noitemsep,topsep=0pt,leftmargin=16pt]
    \item \textbf{Deterministic $A^*$ dataset:} 
    Sequences are generated by executing $A^*$ in a deterministic fashion (by ordering child nodes and breaking equal cost ties deterministically).
    Consequently, given a task prompt, the optimal plan and $A^*$ execution trace is unique.
    Here, the Transformer learns the deterministic breaking rules implicitly encoded in the data.
    Evaluating such a model is simple, because the generated sequences need to exactly match the sequences generated by $A^*$.
    
    \item \textbf{Non-deterministic $A^*$ dataset:}
    Sequences are generated by executing $A^*$ in a non-deterministic fashion (by randomly ordering child nodes and breaking equal cost ties randomly). 
    Consequently, given a task prompt, the optimal plan and $A^*$ execution trace is no longer unique and there are multiple correct responses.
    Here, the Transformer learns to generate the random tie breaking rules implicitly encoded in the sequence data. 
    Consequently, the generated sequences vary between different executions, but the resulting plans are still optimal and execution traces still respect $A^*$'s cost and heuristic calculations as described in Section~\ref{sec:bootstrapping}.
\end{enumerate}

\autoref{fig:sequence-length-stats} in Appendix~\ref{app:dataset-generation} presents an overview of the token sequence length for each dataset and shows that the generated $A^*$ execution traces grow in length with task complexity.
\autoref{fig:sequence-length-distribution-all} shows that training and test sets are matched in difficulty and have comparable trace lengths.
For each task, one model may generate a search sequence ending either in an optimal plan, a feasible plan (a plan that is correct but sub-optimal), or an invalid plan. 
In Appendix~\ref{app:evaluation-criteria} we outline how each model's ability to predict a feasible and optimal plan is scored and details about how the search dynamics of the \plantrace{} and \ours{} models is evaluated.

Unless indicated otherwise, each experiment is repeated five times and each figure plots averages across all repeats. All reported errors indicate the Standard Error of Measurement (SEM).

\subsection{Maze navigation}\label{sec:maze-experiments}

In the first experiment set, we train a set of encoder-decoder Transformer models to predict optimal plans for maze navigation tasks.
We vary the training dataset size and model size (the number of optimized parameters) between different training runs and evaluate each model on the test tasks generated using the same hyper-parameters.

\begin{figure}
    \vspace{-2mm}
    \centering
    \subfigure[
        Deterministic case
    ]{
        \label{fig:maze-trace-added-det}
        \includegraphics[scale=1]{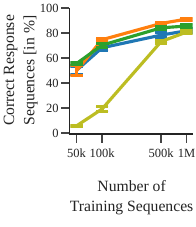}
    }~~~
    \subfigure[
        Non-deterministic case
    ]{
        \label{fig:maze-trace-added-nondet}
        \includegraphics[scale=1]{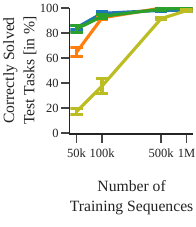}
        \hspace{2mm}
    }~~~
    \subfigure[
        Performance across task difficulties 
    ]{
        \label{fig:maze-size-comparison}
        \includegraphics[scale=1]{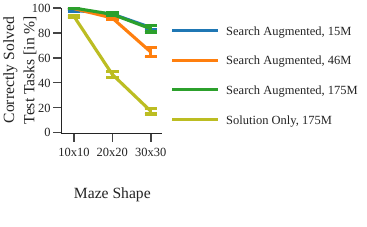}
    }
    \vspace{-4mm}
    \caption{
        \textbf{Predicting intermediate computation steps leads to robust performance in the low data regime.}
        For each model, the number of free parameters (indicated in millions of parameters with ``15M'', ``46M'', and ``175M'') is varied.
        \subref{fig:maze-trace-added-det}: Comparison of how many test tasks were answered with a correct token sequence when training on the deterministic $A^*$ dataset (exact-match criterion in Appendix~\ref{app:evaluation-criteria}).
        \subref{fig:maze-trace-added-nondet}: Comparison for how many test task at least one optimal plan was found when training on the non-deterministic $A^*$ dataset (any-optimal-64 criterion in Appendix~\ref{app:evaluation-criteria}).
        \subref{fig:maze-size-comparison}: Performance degradation when increasing task difficulty (maze size). 
        Here, the non-deterministic $A^*$ dataset was used and models are evaluated with the any-optimal-64 criterion.
    }
    \label{fig:maze-trace-added-results}
\end{figure}

\subsubsection*{Deterministic \texorpdfstring{$A^*$}{A*}}

\autoref{fig:maze-trace-added-det} plots for how many test tasks a correct response was generated. 
Both \planonly{} and \plantrace{} models are trained on the deterministic $A^*$ dataset and are evaluated if they exactly re-produce the token sequences generated with $A^*$ search (please refer to the exact-match criterion in Appendix~\ref{app:evaluation-criteria}).
One can observe that the \planonly{} model is outperformed by most \plantrace{} models.
Only for large enough training datasets, the \planonly{} model matches the performance of the worst \plantrace{} model.
In the low training data regime (100,000 training sequences and less), performance of the \planonly{} model degrades significantly, while the performance of each \plantrace{} model stays relatively high.

This result is surprising, because for more than 90\% of the test mazes, the \plantrace{} models generate \texttt{<trace><plan>}-formatted sequences that are thousands of tokens long without predicting any single token incorrectly.
Moreover, the \planonly{} models, that on average predict ten times shorter sequences, are significantly outperformed by the \plantrace{} models.
Even the smallest \plantrace{} model significantly outperforms the much larger \planonly{} model parameters.

This result highlights the power of training Transformers to generate long algorithm execution traces.
We do not observe compounding prediction errors that usually limit deep model-based RL agents~\citep{asadi2018compoundingerrors}, because the used backward-causal decoder network constructs for an $n$-step sequence an $n \times n$ attention map.
Here, this property of the Transformer architecture is used to boost performance when predicting an optimal plan.

\subsubsection*{Non-deterministic \texorpdfstring{$A^*$}{A*}}

When trained on non-deterministic $A^*$ data, the model could output multiple different optimal paths for one task.
Here, we use each model to generate 64 token sequences for each task.
The test task is counted as correctly answered of any one of the 64 sequences contains an optimal plan (please refer to the any-optimal-64 criterion in Appendix~\ref{app:evaluation-criteria}).
Because we only test if at least one generated sequence contains an optimal plan, we obtain higher absolute numbers in~\autoref{fig:maze-trace-added-nondet} than in~\autoref{fig:maze-trace-added-det}.

\autoref{fig:maze-trace-added-nondet} plots for how many test tasks an optimal plan was found when generating for each test task 64 token sequences.
Here, we can observe a pattern similar to the deterministic $A^*$ dataset: even the smallest \plantrace{} models outperform \planonly{} model, especially for a small training set.
Moreover, we found that model size only impacts the performance of each of the \plantrace{} models when using very small training datasets (50,000 training sequences).
For larger training dataset sizes no significant difference is found.
Increasing the number of parameters of the \planonly{} models does not significantly improve their performance in the low data regime (\autoref{fig:maze-trace-free} in Appendix~\ref{app:experiment-figures}).

\subsubsection*{Performance across different task difficulty levels}

Lastly,~\autoref{fig:maze-size-comparison} illustrates how a task's difficulty influences the performance of each model.
Here, we focus again on the dataset generated by non-deterministic $A^*$, and consider the number of correctly solved test tasks as a function of maze size.
The larger the maze, the larger the task's state space and the more computation is required to find an optimal solution plan.
While the \planonly{} model's performance drops rapidly as the tasks become more challenging, the \plantrace{} models maintain a comparably high accuracy, even for its smallest model size.
Appendix~\ref{app:experiment-figures} presents a full comparison across all maze sizes.

Overall, while the \planonly{} models learn to predict an optimal plan if the used training dataset is large and diverse enough, \plantrace{} models perform significantly better in the low data regime and scale better to more difficult tasks.
The \plantrace{} models reach higher performance because they can perform on-demand computation during inference.
More specifically, the \plantrace{} models imitate the search dynamics for a grounded reasoning chain that leads to an optimal plan, while the \planonly{} models have to infer direct correlations between a task description and its optimal plan through supervised learning where many of such correlations can be spurious and unreliable during evaluation on the test task set.

\subsection{Solving Sokoban puzzles}\label{sec:sokoban-experiments}

\begin{table}
    \caption{
        \textbf{Test set performance in the Sokoban tasks.} 
        Over 200 unseen test Sokoban tasks, we report percentage of solved and optimally solved test tasks.
        For sequences ending in either an optimal and correct plan we report the SWC (\emph{Success Weighted by Cost}) score, and ILR (\emph{Improved Length Ratio of Search Dynamics}) scores. The better trace and solution quality, the higher the scores.
        Please check Appendix~\ref{app:evaluation-criteria} for detailed definitions of these scores. 
    }
    \begin{center}
        \fontsize{8}{10.5}\selectfont
        \begin{tabular}{r l c c c c c}
            \hline
            Params. & Model                             & Solved {\scriptsize(\%)}     & Optimal {\scriptsize(\%)}    & SWC ($\uparrow$)                            & ILR (solved, $\uparrow$)                     & ILR (optimal, $\uparrow$)                    \\
            \hline \hline
            \multirow{5}{*}{45M} & \PlanonlyNoHyphen{}  & 90.3 {\scriptsize $\pm 1.0$} & 86.8 {\scriptsize $\pm 0.3$} & 0.890 {\scriptsize $\pm 0.009$} & --                              & --                              \\
            & \PlantraceNoHyphen{}                      & 92.5 {\scriptsize $\pm 1.0$} & 90.8 {\scriptsize $\pm 1.6$} & 0.924 {\scriptsize $\pm 0.011$} & 0.908 {\scriptsize $\pm 0.020$} & 0.919 {\scriptsize $\pm 0.019$} \\
            & \planimproved{}, step 1                   & 95.5 {\scriptsize $\pm 1.0$} & 93.5 {\scriptsize $\pm 1.0$} & 0.953 {\scriptsize $\pm 0.010$} & 1.054 {\scriptsize $\pm 0.025$} & 1.062 {\scriptsize $\pm 0.015$} \\
            & \planimproved{}, step 2                   & 96.0 {\scriptsize $\pm 0.5$} & 93.4 {\scriptsize $\pm 0.6$} & 0.957 {\scriptsize $\pm 0.005$} & 1.158 {\scriptsize $\pm 0.025$} & 1.181 {\scriptsize $\pm 0.012$} \\
            & \planimproved{}, step 3                   & 95.5 {\scriptsize $\pm 0.8$} & 93.7 {\scriptsize $\pm 1.6$} & 0.953 {\scriptsize $\pm 0.009$} & 1.292 {\scriptsize $\pm 0.044$} & 1.343 {\scriptsize $\pm 0.067$} \\
            \hline
            \multirow{2}{*}{175M} & \PlanonlyNoHyphen{} & 95.7 {\scriptsize $\pm 0.2$} & 90.0 {\scriptsize $\pm 0.8$} & 0.949 {\scriptsize $\pm 0.003$} & --                              & --                              \\
            & \PlantraceNoHyphen{}                      & 95.2 {\scriptsize $\pm 0.9$} & 93.2 {\scriptsize $\pm 1.0$} & 0.949 {\scriptsize $\pm 0.010$} & 0.925 {\scriptsize $\pm 0.010$} & 0.933 {\scriptsize $\pm 0.011$} \\
            \hline
            757M & \PlanonlyNoHyphen{}                  & 96.5 {\scriptsize $\pm 0.1$} & 92.2 {\scriptsize $\pm 1.2$} & 0.958 {\scriptsize $\pm 0.002$} & --                              & --                              \\
            \hline
        \end{tabular}
    \end{center}
    \label{tab:performance}
    \vspace{-2mm}
\end{table}

To test if similar results can be obtained on a different and more complex task with a different tokenization pattern and different transition structure, we repeat our experiments for Sokoban puzzles using our non-deterministic $A^*$ implementation.
\autoref{tab:performance} lists how often each model generated a correct optimal plan for each test task.
As before, by training on execution traces, the \plantrace{} models outperform the \planonly{} models.
Even increasing the parameterization of a \planonly{} model to 747 million parameters only leads to a marginal performance improvement.
On average, this 747 million parameter \planonly{} model is still outperformed slightly by a smaller 175 million parameter \plantrace{} model.
This experiment further confirms our findings on more complex planning tasks with a different transition structure and a different tokenization method.

\subsection{\ours{}: Improving search dynamics via bootstrapping}
\label{sec:trace-improvement}

In this last experiment, we investigate how the \plantrace{} model can be iteratively improved to compute an optimal plan while generating a shorter execution trace.
Here, our goal is to shorten the length of the search trace while still producing an optimal solution.

We start out with the smallest \plantrace{} model trained on the non-deterministic $A^*$ Sokoban dataset and use it to generate a new shorter sequence training dataset as outlined in Section~\ref{sec:bootstrapping}.
For each Sokoban puzzle in the training data, we generated 32 different \texttt{<trace><plan>}-formatted sequences by sampling tokens from the Transformer's output distribution and include the shortest generation (measured in tokens) if it contains an optimal plan.
Subsequently, we fine-tune the \plantrace{} model on this newly created training data (by running an additional 10,000 training steps) to obtain the first \planimproved{} model.
Using this \planimproved{} model, we subsequently generate another short sequence dataset and repeat the fine-tuning procedure to further improve the model.

\begin{figure}
    \centering
    \subfigure[
        Search length improvement.
    ]{
        \label{fig:searchformer-avg-trace-len}
        \includegraphics[scale=1]{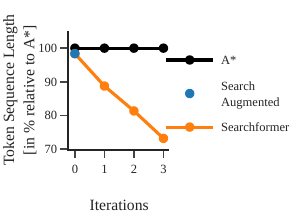}
    }~~~
    \subfigure[
        Distribution of average-on-optimal length.
    ]{
        \label{fig:searchformer-hist}
        \includegraphics[scale=1]{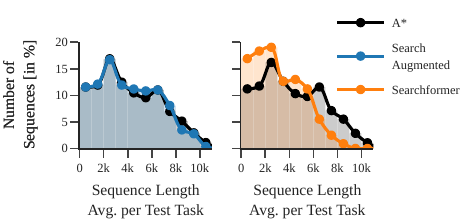}
    }
    \vspace{-4mm}
    \caption{
        \textbf{Improvement of search dynamics length via bootstrapping in Sokoban}
        \subref{fig:searchformer-avg-trace-len}: 
        For each Sokoban test task that was answered with an optimal plan, the average generated execution trace length is computed and averaged.
        The $A^*$ reference values are computed by generating an equal number of execution traces for each test task. 
        For each iteration, we compare the subset of test tasks that were answered with an optimal plan by the search-augmented and Searchformer model and compare the relative improvement over our $A^*$ reference implementation.
        \autoref{fig:searchformer-box-plot} (Appendix~\ref{app:experiment-figures}) shows a box plot with more details about the sequence length distribution.
        \subref{fig:searchformer-hist}:
        Distribution of execution trace lengths for each optimally answered test task.
        Here, the sequence lengths were first averaged per test task and the resulting distribution is plotted in each histogram.
        The \plantrace{} model is trained to imitate our $A^*$ implementation.
        After three iterations of search dynamics bootstrapping, the resulting \ours{} model generates on average shorter execution traces.
    }
    \label{fig:searchformer}
\end{figure}

\autoref{fig:searchformer-avg-trace-len} illustrates how the sequence lengths generated by the \planimproved{} model's are iteratively shortened with our search dynamics boostrapping method.
With every improvement step, the average length of the generated traces---the number of search steps---decreases (\autoref{fig:searchformer-avg-trace-len}).
When computing an optimal plan, the final \planimproved{} model generates search dynamics sequences that are on average 26.8\% shorter than the sequences generated with $A^*$ search.
Consequently, the \planimproved{} model found a way to find a plan in a complex task that is more efficient in terms of search steps than the $A^*$ implementation used to train the initial \plantrace{} model.
In~\autoref{fig:searchformer-hist} we can observe that the \plantrace{} model generates sequences that on average match the sequences generated with $A^*$ search in length.
The \planimproved{} models generate shorter sequences resulting in a distribution that is skewed towards shorter sequence lengths.

As reported in \autoref{tab:performance}, fine-tuning the model resulted in a significant performance improvement, reducing the rate of incorrect and non-optimal solutions by 40\% and 30\% respectively. 
The \emph{Success Weighted by Cost} (SWC) score~\citep{wu2019bayesianrelationalmemory} factors in how many test tasks were solved correctly and how close the predicted plans are to the optimal length (Appendix~\ref{app:evaluation-criteria}).
Here, a perfect score would be one, and one can see in~\autoref{tab:performance} that the comparably small \ours{} matches the performance of the largest \planonly{} model (also note the small SEM values).
Furthermore, the \emph{Improved Length Ratio of Search Dynamics} (ILR) measures how much the length of each execution trace is shortened (Appendix~\ref{app:evaluation-criteria}). 
With each improvement iteration the scores increase and climb above one.
For example, $A^*$ search dynamics is $\sim 34.3\%$ longer than the sequences generated by \planimproved{} after 3 steps of fine-tuning. 

The results reported in~\autoref{fig:searchformer} and in~\autoref{tab:performance} only compare each model's performance on test tasks that were either correctly or optimally solved.
To test if a model overfits only on easier test tasks with shorter execution traces, we plot in~\autoref{fig:searchformer-scatter} (in Appendix~\ref{app:searchformer-performance}) the execution trace length generated with $A^*$ against the execution trace length generated by each model as a scatter plot.
Each point in this plot corresponds to a single test task.
Here, the trend of shortening the execution trace via search dynamics bootstrapping is clear and one can also observe that neither model only specializes on solving easier test tasks with shorter execution traces.

\section{Discussion}

Prior work~\citep{momennejad2023evaluatingllms,valmeekam2023planningstudy} has found that LLMs struggle with solving complex decision making tasks.
\ours{} demonstrates that with appropriate training data, Transformers can in fact solve complex planning tasks.
Moreover, \ours{} robustly follows the intermediate steps---the execution trace---of a symbolic planner and improves (in terms of trace length) beyond the human-crafted rule-based planning strategy it was initially trained on. 
Compared to \planonly{} models that directly predict a solution, our \plantrace{} models require fewer training sequences and scale better to more complex planning tasks.

\subsection{Limitations}

Currently, \ours{} is trained on the execution traces of $A^*$ to learn a complex planning strategy. 
However, the trace length may grow exponentially in the length of an optimal plan (see~\autoref{fig:sequence-length-stats}), and training on the resulting token sequence data can become computationally very expensive. 
In fact, the presented experiments use token sequences that are significantly longer than the sequences used to train LLMs such as Llama 2~\citep{touvron2023llama2}.

\subsection{Future work}

One way to mitigate this limitation and improve the efficiency of the presented methods is to use curriculum learning: starting from simple tasks with reasonably long execution traces, train and fine-tune the \ours{} to reduce the trace length, and then adapt the improved model to more complex tasks. 
Another possibility is to explore other planning algorithms or integrate better heuristics or value functions into $A^*$ search, similar to MCTS, to cap the maximal depth the search algorithm explores. 
Integrating hierarchical planning methods and temporal abstractions~\citep{sutton2023temporalabstractionplanning,sutton1999options,dietterich2000hierarchical,hafner2022hierarchicalplanning} are another avenue. 
This would equip the resulting model with the ability to abstract over multiple time steps and states to find an optimal plan using fewer computation steps.

In comparison to Plansformer~\citep{pallagani2022plansformer}, the presented work demonstrates how to train Transformers from scratch to solve complex planning tasks on synthetic datasets.
We believe that our results and methods could be combined with methods such as Plansformer to fine-tune LLMs and enable them to solve complex planning tasks more robustly.
Ultimately, we hope that our study sheds light on how Transformers can be used for multi-step planning and we hope to inform further research about better understanding the reasoning capabilities of LLMs.

\section{Broader Impact}

Our work focuses on symbolic planning tasks and uses synthetic datasets for training.
While the tasks we explored in this paper can be easily solved with simple symbolic solvers, it is important to study the effectiveness of neural networks on such tasks.
Here, we provide a proof of concept on how Transformer-based neural networks can be used to robustly solve complex planning tasks.
With our work, we hope to inform further research into better understanding the reasoning capabilities of Large Language Models.

\subsection*{Acknowledgment}
We would also like to thank Amy Zhang for helpful discussions on this work.

\newpage
\bibliography{library}
\bibliographystyle{colm/colm2024_conference}

\newpage
\appendix

\section{Training encoder-decoder Transformers to predict optimal plans}
\label{app:methods}

For this work, we consider encoder-decoder Transformers~\citep{raffel2020t5} and process each prompt---the task specification---with an encoder network to obtain an embedding of a planning task.
Using the decoder, the resulting embedding is then decoded into a response of the format \texttt{<trace><plan>} or the format \texttt{<plan>}.
We refer to a model predicting sequences of the format \texttt{<trace><plan>} as a \emph{\plantrace{}} model and a model predicting sequences of the format \texttt{<plan>} as \planonly{} models.

\subsection{Encoder-decoder architecture}
A token sequence is processed by a neural network by first indexing the set of all tokens contained in a specific dataset.
This indexing is used to map any token sequence to a set of integers.
Formally, we denote a prompt as a sequence of integers $x_{1:n} = (x_1,...,x_n)$.
An encoder Transformer neural network with network weights $\pmb{\theta}_\text{enc}$ is a function $f_{ \pmb{\theta}_\text{enc} }$ mapping a tokenized prompt $x_{1:n}$ of arbitrary length $n$ to a sequence of $n$ real-valued vectors:
\begin{equation}
	f_{ \pmb{\theta}_\text{enc} } : x_{1:n} \mapsto \pmb{z}_{1:n}
\end{equation}
Each vector $\pmb{z}_i$ in this sequence of vectors $\pmb{z}_{1:n}$ has the same dimension.
The decoder network is then used to generate a response auto-regressively:
Starting with a specific beginning-of-sequence token \texttt{bos} to cue the decoder, a sequence is recursively built by first mapping the start token \texttt{bos} to a probability distribution over next-tokens.
This probability distribution is stored as a vector $\pmb{p}_1$ whose dimension is equal to the size of the vocabulary.
The next token is then generated by sampling from this distribution and the sampled token is appended to the response sequence.
Subsequently the two-token response sequence is fed into the decoder again to compute the next probability vector $\pmb{p}_2$ and sample the next token.
This procedure is repeated until an end-of-sequence token \texttt{eos} is sampled.
While only the last computed probability vector is needed to sample the next token, the decoder network simultaneously predicts a sequence of next token probability vectors $\pmb{p}_{1:m}$ given an input sequence $y_{1:m}$.
Furthermore, this prediction is conditioned on the encoder output $\pmb{z}_{1:n}$.
The decoder Transformer neural network with weight parameters $\pmb{\theta}_\text{dec}$ is therefore a function
\begin{equation}
	g_{ \pmb{\theta}_\text{dec} }: \pmb{z}_{1:n}, y_{1:m} \mapsto \pmb{p}_{1:m}.
\end{equation}
The encoder network $f_{ \pmb{\theta}_\text{enc} }$ and decoder network $g_{ \pmb{\theta}_\text{dec} }$ internally both use a number of stacked causal attention layers to form an encoder-decoder Transformer~\citep{vaswani2017transformer} as outlined in more detail in Appendix~\ref{app:hyper-parameters}.
We denote a concatenation of all encoder parameters $\pmb{\theta}_\text{enc}$ and decoder parameters $\pmb{\theta}_\text{dec}$ with the vector $\pmb{\theta}$.

\subsection{Training with teacher forcing}
An encoder-decoder architecture is optimized to generate responses that follow the distribution of a training dataset by minimizing the cross-entropy between the training data distribution $p_\mathcal{D}$ over prompt-response pairs $(x_n,y_m)$ and the distribution $p_{\pmb{\theta}}$ with which the encoder-decoder model is generating responses.
This cross-entropy loss objective
\begin{align}
H(p_\mathcal{D},p_{\pmb{\theta}}) &= \mathbb{E}_{\mathcal{D}} \left[ - \log p_{\pmb{\theta}}( y_{1:m} | x_{1:n} ) \right]  \label{eq:cross-ent-loss-1} \\
&= \mathbb{E}_{\mathcal{D}} \left[ -\sum_{i=1}^{m-1} \log p_{\pmb{\theta}}( y_{i+1:m} | y_{1:i}, x_{1:n} ) \right] \label{eq:cross-ent-loss-2},
\end{align} 
where line~\eqref{eq:cross-ent-loss-2} follows from the auto-regressive generation procedure described before.
Within the same planning dataset, different prompt-response pairs can have different prompt and response lengths.
To emphasize shorter response sequences during training, we re-weigh each sample resulting in the loss objective
\begin{equation}
L(\pmb{\theta}) = \frac{1}{D} \sum_{d=1}^D \frac{1}{m_d-1} \sum_{i=1}^{m_d-1} \log p_{\pmb{\theta}}( y^{d}_{i+1:m_d} | y^{d}_{1:i}, x^{d}_{1:n_d} ), \label{eq:loss-objective}
\end{equation}
where the first summation averages over all $D$ prompt-response pairs of the training dataset.
In Equation~\eqref{eq:loss-objective} the super-script $d$ indexes individual prompt-response pairs in the training dataset.
This average is an empirical approximation of the expectation in Equation~\eqref{eq:cross-ent-loss-1} for a finite i.i.d. sample of size $D$.
This loss objective is optimized using gradient descent~\citep[Chapter 10]{goodfellow2016deeplearning}.

\begin{figure}[h!]
    \centering
    \includegraphics[width=\textwidth]{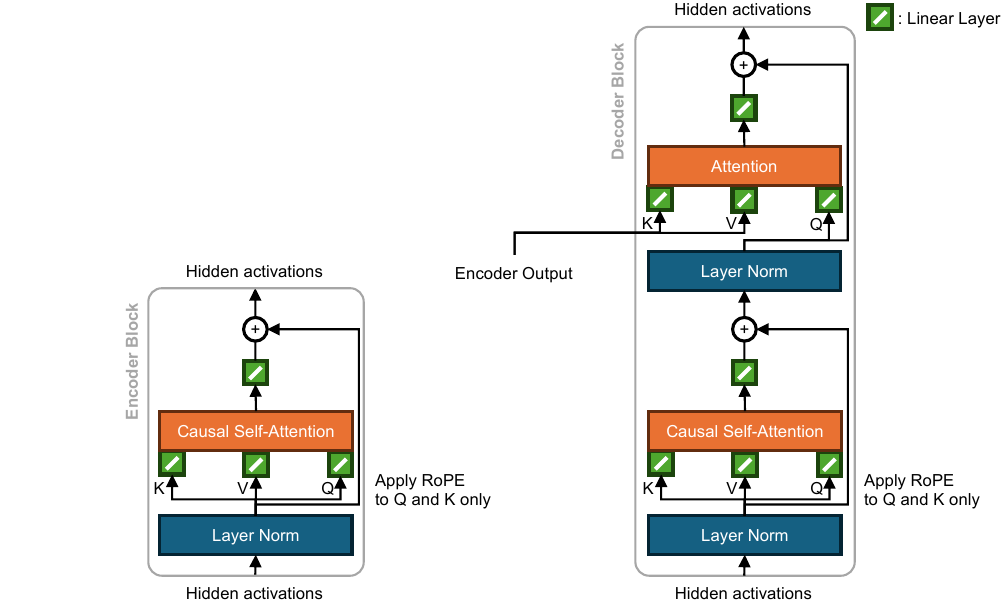}
    \caption{
        \textbf{Attention blocks architecture used in the encoder-decoder Transformer architecture.}
        The encoder network consists of a set of feed-forward layers where each layer processes hidden activations in parallel with a set of encoder attention blocks (left diagram).
        Similarly, the decoder network consists of a set of feed-forward layers composed of a number of decoder attention blocks (right diagram).
        The number of blocks in each layer is referred to as the number of heads.
        Token sequences are first mapped into integer sequences using a look-up dictionary.
        Then, these sequences are fed through a PyTorch~\citep{paszke2019pytorch} \texttt{torch.nn.Embedding} layer to map the integer sequence into a sequence of hidden activation vectors.
        After the last decoder layer, hidden activations are mapped through a linear layer into a sequences of logits vectors to compute the next token probability vectors $\pmb{p}_{1:m}$.
    }
    \label{fig:architecture}
\end{figure}

\section{Network architecture and hyper-parameters}
\label{app:hyper-parameters}

The encoder-decoder Transformer first maps every token to a one-hot vector of the same dimension as the token vocabulary space.
These one-hot vectors are then projected through a linear embedding layer into a set of vectors.

The encoder then consists of multiple feed-forward layers and each layer consists of multiple encoder blocks (left part of~\autoref{fig:architecture}).
The output of these layers is then mapped through another linear layer to project the hidden activations into a tensor of the correct shape.
The decoder also consists of multiple feed-forward layers and each layer also consists of multiple decoder blocks (right part of~\autoref{fig:architecture}) processing the hidden activations in parallel.
As illustrated in~\autoref{fig:architecture}, the decoder network is conditioned on the encoder by processing the encoder's output tensor directly in the second attention map.
Furthermore, each tokens position is encoded by applying RoPE embeddings~\citep{su2023rope} as indicated in~\autoref{fig:architecture}.
We did not use dropout in our architecture.

\begin{table}[h!]
    \caption{
        \textbf{Architecture Hyper-parameters.}
        The same parameters were used in both the encoder and decoder network.
        The number of heads indicates how many attention blocks are used in one layer.
        Layer dimension indicates the dimension of the feature vectors processed through each attention block (dimension of $K$, $V$, and $Q$ in~\autoref{fig:architecture}).
        All models used a RoPE frequency of 10000.
    }
    \label{tab:architecture-params}
    \begin{center}
        \begin{small}
            \begin{tabular}{lcccc}
                \hline
                Parameter  & 15M model  & 46M model  & 175M model & 747M  model \\
                \hline
                Layers     & 6                 & 8                 & 9                 & 16                 \\
                Heads      & 3                 & 4                 & 4                 & 12                 \\
                Layer dim. & 64                & 96                & 192               & 96                 \\
                \hline
            \end{tabular}
        \end{small}
    \end{center}
\end{table}

\begin{table}[h!]
    \caption{
        \textbf{Optimization hyper-parameters.}
        Every model was optimized using AdamW~\citep{loshchilov2019adamw} with setting $\beta_0=0.9$ and $\beta_1=0.99$.
        Initially, the learning rate was linearly interpolated: Starting at zero and then increasing linearly to the value listed below until step 2000. 
        Then a cosine learning rate schedule was followed~\citep{loshchilov2016cosinelr}.
        to the between zero at the first training step and the listed value below for the 
    }
    \label{tab:opt-param}
    \begin{center}
        \begin{small}
            \begin{tabular}{llcc}
                \hline
                Parameter      & Model & Maze Tasks          & Sokoban Puzzels     \\
                \hline \hline
                Learning rate  &  15M  & $2.5 \cdot 10^{-4}$ & $2.5 \cdot 10^{-4}$ \\
                               &  46M  & $7.5 \cdot 10^{-5}$ & $7.5 \cdot 10^{-5}$ \\
                               & 175M  & $5.0 \cdot 10^{-5}$ & $5.0 \cdot 10^{-5}$ \\
                               & 747M  & --                  & $5.0 \cdot 10^{-5}$ \\
                \hline
                Batch size     & all   & 16                  & 64                  \\
                \hline
                Training steps & all   & 400000              & 80000               \\
                \hline
            \end{tabular}
        \end{small}
    \end{center}
\end{table}

\autoref{tab:architecture-params} lists the used architecture hyper-parameter and~\autoref{tab:opt-param} lists the hyper-parameters used for optimizing each model.
All experiments were implemented in PyTorch 2.0~\citep{paszke2019pytorch} and default parameters were used unless reported here otherwise.

\newpage 

\section{Dataset generation}
\label{app:dataset-generation}

All datasets were generated by first randomly sampling a task and then executing $A^*$ to obtain an optimal plan.
Maze tasks were generated first by randomly selecting 30-50\% of all cells to be wall cells.
Then a start and goal location was randomly selected and $A^*$ was executed to obtain an optimal plan.
If the plan had a length of at least the mazes width or height (e.g. for $10 \times 10$ mazes the optimal plan needs to contain at least 10 steps), then the task was added into the dataset.
For Sokoban, a $7 \times 7$ grid was sampled and two additional wall cells were added as obstacles to the interior of the map. 
Then two docks, boxes, and the worker locations were randomly placed.
If the sampled task is solvable by A*, then the task was admitted to the dataset.

\begin{wrapfigure}{r}{0.45\textwidth}
    \centering
    \includegraphics[height=3cm]{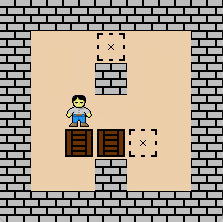}
    \caption[]{
        \textbf{Example Sokoban puzzle\footnotemark.}
        The prompt, $A^*$ execution trace, and optimal plan for this task is illustrated in ~\autoref{fig:sokoban-trace} in Appendix~\ref{app:dataset-generation}.
        For Sokoban, we only use a non-deterministic $A^*$ implementation.
    }
    \label{fig:sokoban-level}
\end{wrapfigure}

\begin{figure}[h!]
    \centering
    \subfigure[Plan Sequence Length]{
        \label{fig:plan-length-distribution-all}
        \includegraphics[scale=0.9]{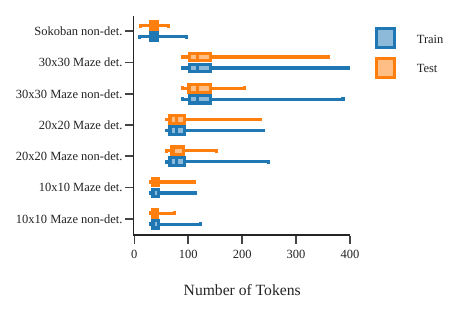}
    }
    \subfigure[$A^*$ Execution Trace Length]{
        \label{fig:execution-trace-jength-distribution-all}
        \includegraphics[scale=0.9]{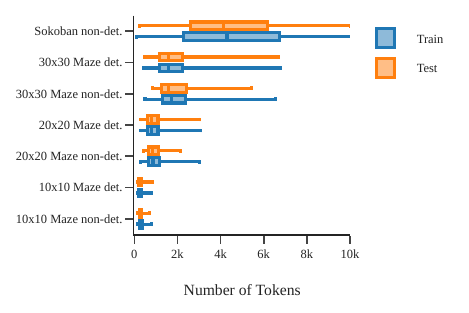}
    }
    \caption{
        \textbf{Sequence length distribution for each dataset.}
        The training and test sets are designed such that their sequence lengths match closely.
    }
    \label{fig:sequence-length-distribution-all}
\end{figure}

Due to the large volume of generated data, all data was stored in and transformed with a MongoDB~\citep{mongodb} instance and map-reduce techniques.
Furthermore, when sampling tasks, the dataset was constructed to reject duplicate tasks ensuring that training and test data and each prompt is distinct.
Once each task and trace dataset was generated, each execution trace is converted into prompt and response token sequences, as illustrated in~\autoref{fig:astar-trace} and~\autoref{fig:sequence-length-distribution-all}.
Because the Sokoban tasks are very complex and difficult to solve for A*, the resulting token sequences were partially very long and reached almost 100000 tokens.
Due to GPU memory requirements, the Sokoban dataset was further sliced to only include sequences of with at most 10000 tokens.
\autoref{fig:sequence-length-distribution-all} compares the sequence length distribution for each dataset.
During training, each dataset was also sorted and sliced to only contains the reported number of training sequences.
Furthermore, each model was evaluated on the same test tasks within each task dataset.
The test dataset contains plans and traces that are of comparable length to the training data (\autoref{fig:sequence-length-distribution-all}).

\footnotetext{\footnotesize This level image was composed using image icons from \url{https://github.com/morenod/sokoban} (accessed 2023-11-21).}

\begin{figure}[h!]
    \centering
    \renewcommand{\ttdefault}{pcr}
\renewcommand{\sfdefault}{cmr}
\begin{tikzpicture}
    \node[align=left,anchor=west,font={\scriptsize\ttfamily}] at (-3.8,6.0) {bos};  
    \node[align=left,anchor=west,font={\scriptsize\ttfamily}] at (-3.8,5.7) {worker 2 3};
    \node[align=left,anchor=west,font={\scriptsize\ttfamily}] at (-3.8,5.4) {box \ \ \ 2 4};
    \node[align=left,anchor=west,font={\scriptsize\ttfamily}] at (-3.8,5.1) {box \ \ \ 3 4};
    \node[align=left,anchor=west,font={\scriptsize\ttfamily}] at (-3.8,4.8) {dock \ \ 1 3};
    \node[align=left,anchor=west,font={\scriptsize\ttfamily}] at (-3.8,4.4) {dock \ \ 4 4};
    \node[align=left,anchor=west,font={\scriptsize\ttfamily}] at (-3.8,4.0) {wall \ \ 0 0};
    \node[align=left,anchor=west,font={\scriptsize\ttfamily}] at (-3.8,3.7) {wall \ \ 0 1};
    \node[align=left,anchor=west,font={\scriptsize\ttfamily}] at (-3.8,3.4) {...};
    \node[align=left,anchor=west,font={\scriptsize\ttfamily}] at (-3.8,3.1) {wall \ \ 6 6};
    \node[align=left,anchor=west,font={\scriptsize\ttfamily}] at (-3.8,2.8) {eos};
    
    \node[align=right,anchor=east,font={\scriptsize}] at (-4.0, 4.3) {Prompt};	
    \draw[looseness=0.4] (-4.0, 4.3) to[out=0,in=-180]  (-3.8,6.15);
    \draw[looseness=0.4] (-4.0, 4.3) to[out=0,in=-180]  (-3.8,2.75);

    \node[align=left,anchor=west,font={\scriptsize\ttfamily}] at (0,6.0) {bos};
    \node[align=left,anchor=west,font={\scriptsize\ttfamily}] at (0,5.7) {create worker 2 3 box 2 4 box 3 4 c0 \ c3};
    \node[align=left,anchor=west,font={\scriptsize\ttfamily}] at (0,5.4) {close \ worker 2 3 box 2 4 box 3 4 c0 \ c3};
    \node[align=left,anchor=west,font={\scriptsize\ttfamily}] at (0,5.1) {...};
    \node[align=left,anchor=west,font={\scriptsize\ttfamily}] at (0,4.8) {create worker 5 4 box 2 3 \ \ \ \ \ \ \ \ c10 c3};
    \node[align=left,anchor=west,font={\scriptsize\ttfamily}] at (0,4.5) {close \ worker 2 1 \ \ \ \ \ \ \ \ \ \ \ \ \ \ \ \ c12 c0};
    
    \node[align=right,anchor=east,font={\scriptsize}] at (-0.2, 5.25) {Trace\\(2583 tokens)};	
    \draw[looseness=0.4] (-0.2, 5.25) to[out=0,in=-180]  (0.0,6.15);
    \draw[looseness=0.4] (-0.2, 5.25) to[out=0,in=-180]  (0.0,4.35);
    
    \node[align=left,anchor=west,font={\scriptsize\ttfamily}] at (7.5,6.0) {plan 2 3};
    \node[align=left,anchor=west,font={\scriptsize\ttfamily}] at (7.5,5.7) {plan 1 3};
    \node[align=left,anchor=west,font={\scriptsize\ttfamily}] at (7.5,5.4) {plan 1 4};
    \node[align=left,anchor=west,font={\scriptsize\ttfamily}] at (7.5,5.1) {plan 1 5};
    \node[align=left,anchor=west,font={\scriptsize\ttfamily}] at (7.5,4.8) {plan 2 5};
    \node[align=left,anchor=west,font={\scriptsize\ttfamily}] at (7.5,4.5) {plan 2 4};
    \node[align=left,anchor=west,font={\scriptsize\ttfamily}] at (7.5,4.2) {plan 3 4};
    \node[align=left,anchor=west,font={\scriptsize\ttfamily}] at (7.5,3.9) {plan 2 4};
    \node[align=left,anchor=west,font={\scriptsize\ttfamily}] at (7.5,3.6) {plan 2 3};
    \node[align=left,anchor=west,font={\scriptsize\ttfamily}] at (7.5,3.3) {plan 2 2};
    \node[align=left,anchor=west,font={\scriptsize\ttfamily}] at (7.5,3.0) {plan 1 2};
    \node[align=left,anchor=west,font={\scriptsize\ttfamily}] at (7.5,2.7) {plan 1 1};
    \node[align=left,anchor=west,font={\scriptsize\ttfamily}] at (7.5,2.4) {plan 2 1};
    \node[align=left,anchor=west,font={\scriptsize\ttfamily}] at (7.5,2.1) {eos};
    
    \node[align=right,anchor=east,font={\scriptsize}] at (7.3, 4.05) {Plan};	
    \draw[looseness=0.4] (7.3, 4.05) to[out=0,in=-180]  (7.5,6.15);
    \draw[looseness=0.4] (7.3, 4.05) to[out=0,in=-180]  (7.5,1.95);

\end{tikzpicture}
    \caption{
        \textbf{Token sequence example for Sokoban}
        This figure lists the token sequence for the Sokoban level depicted in~\autoref{fig:sokoban-level}.
    }
    \label{fig:sokoban-trace}
\end{figure}

\begin{figure}[h!]
    \centering
    \includegraphics[scale=1.0]{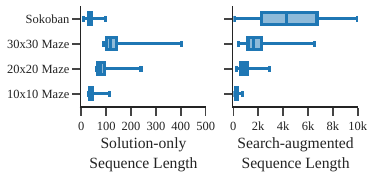}
    \caption{
        \textbf{Training Sequence Length Comparison.}
        The left panel plots the length of the \planonly{} sequences and the right panel plots the length of the \plantrace{} sequences, excluding the start and end of sequence tokens \texttt{bos} and \texttt{eos}.
        The whiskers indicate the range of all sequence lengths and the box plots indicate the 25\%, 50\%, and 75\% quantiles.
        Because we focus on planning in complex sequential decision making tasks, the token sequences are multiple orders of magnitude longer than usual token sequences used to train LLMs---especially when $A^*$ execution traces are included in the responses.
        For example, fine-tuning of the Llama 2 model on human preference data is performed with sequences that are on average 600 tokens long~\citep{touvron2023llama2}.
    }
    \label{fig:sequence-length-stats}
\end{figure}

\newpage

\section{Evaluation criteria}\label{app:evaluation-criteria}

In the presented experiments we evaluate if each model outputs an optimal or feasible plan and how long the generated search sequences are for the \plantrace{} and \ours{} models.

\subsection{Measuring plan optimality}
\label{sec:eval-plan-optimality}
We evaluate whether the output plan is optimal using one of three criteria:
\begin{itemize}[noitemsep,topsep=0pt,leftmargin=16pt]
    \item \textbf{Exact-match criterion.} 
    For each task, if the generated sequence from a trained model matches the output of deterministic $A^*$ exactly, it is labelled as correct, otherwise labelled as incorrect. 
    This is only used to evaluate supervised cloning of deterministic $A^*$. 
    \item \textbf{Any-optimal-64 criterion.} 
    For each task, we sample 64 responses from a trained model. 
    Each response is parsed and evaluated if it contains a feasible or optimal plan, regardless of the generated \texttt{<trace>} part. 
    If any of the 64 plans is feasible and optimal, then the task is labelled as correct. 
    \item \textbf{SWC score.} 
    To further measure the sub-optimalness of the resulting plans, we also report the \emph{Success Weighted by Cost (SWC)}~\citep{wu2019bayesianrelationalmemory}, a statistic that factors in how close the cost $l_i$ of the best predicted correct plan (over 64 trials) is to the optimal plan cost $l_i^*$, averaged across all $n$ test tasks and weighted by a binary variable $c_i \in \{ 0, 1 \}$:
    \begin{equation*}
        \text{SWC} := \frac{1}{n} \sum_{i=1}^n c_i \frac{l_i^*}{\max \{ l_i,l_i^* \} } .
    \end{equation*}
    When computing the SWC score, the binary variable $c_i$ is set to one if a correct plan was found and zero otherwise.
    This SWC score lies between zero and one. If all generated sequences end in an optimal plan, then this value is one. 
\end{itemize}

\subsection{Measuring search dynamics length}
\label{sec:eval-search-dynamic-length}
For sequences ending in an optimal or feasible plan, we evaluate the length of sequence dynamics in terms of number of tokens using one of the two criteria:

\begin{itemize}[noitemsep,topsep=0pt,leftmargin=16pt]
    \item \textbf{Average-on-optimal length.} 
    For each task, we sample 64 responses from a trained model and compute averaged search dynamics length for sequences that lead to an optimal plan.
    \item \textbf{ILR score.} 
    To further measure how much improvement of the model-generated search dynamics against $A^*$ planning, we report the \emph{Improved Length Ratio of Search Dynamics} (ILR) score.
    Specifically, for each test task $i$, we compute the ratio between the length $t_i$ of the shortest generated search dynamics sequence and the $A^*$ token sequence length $t_i^*$. 
    We then average across all test tasks while only including ratios for tasks for which either an optimal or a correct (and potentially sub-optimal) plan was found, as specified by the binary variable $c_i$. 
    The corresponding measures are thus called \emph{ILR-on-optimal} and \emph{ILR-on-solved}. 
    The ILR is defined as
    \begin{equation*}
        \text{ILR} := \frac{1}{n} \sum_{i=1}^n c_i \frac{t_i^*}{t_i}.
    \end{equation*}
    The ILR measure can take non-negative values and values above one indicate that the model generates shorter search dynamics than the $A^*$ reference. Consequently, if the numbers lie significantly above one, then the model has found a more efficient way to search a task's state space to compute an optimal plan.
\end{itemize}


\section{\ours{} performance analysis}\label{app:searchformer-performance}

\begin{figure}[h!]
    \centering
    \subfigure[Comparison of all sequences ending with an optimal plan]{
        \label{fig:searchformer-scatter-optimal} 
        \includegraphics[scale=1]{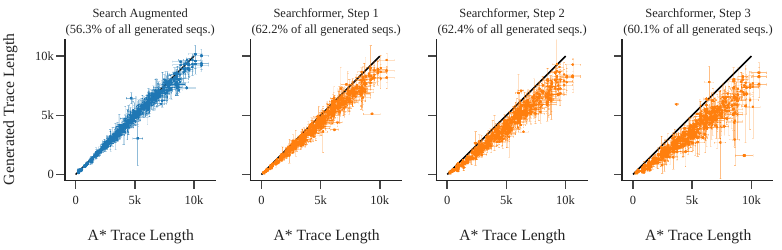}
    }
    \subfigure[Comparison of all sequences ending with a correct plan]{
        \label{fig:searchformer-scatter-correct} 
        \includegraphics[scale=1]{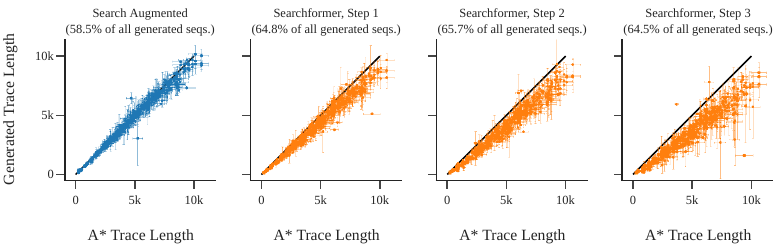}
    }
    \subfigure[Comparison of all sequences ending with a correct but sub-optimal plan]{
        \label{fig:searchformer-scatter-correct-but-suboptimal} 
        \includegraphics[scale=1]{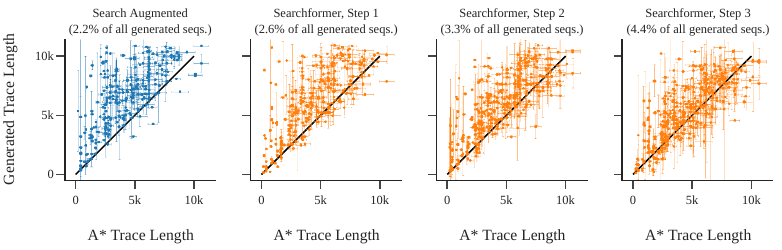}
    }
    \vspace{-2mm}
    \caption{
        \textbf{Sequence length comparison between sequences generated with $A^*$ search and sequences generated with each model.}
        Each dot in each scatter plot corresponds to one specific test task.
        On the $x$-axis, we plot the average token sequence length when $A^*$ search is used.
        On the $y$-axis, we plot the average token sequence length when each model is used.
        Error bars indicate standard deviations.
        Percentages indicate the fraction of the generated sequences that are included in each scatter plot.
        \subref{fig:searchformer-scatter-optimal}: Sequence length comparison for all test prompts for which an optimal plan was generated.
        \subref{fig:searchformer-scatter-correct}: Sequence length comparison for all test prompts for which a correct plan was generated.
        This plot aggregates across sequences ending in an optimal plan and sequences ending in a correct but sub-optimal plan.
        \subref{fig:searchformer-scatter-correct-but-suboptimal}: Sequence length comparison for all test prompts for which a correct but sub-optimal plan was generated using the corresponding model.
        This plot only aggregates across sequences ending in a correct but sub-optimal plan.
    }
    \label{fig:searchformer-scatter}
\end{figure}

In Section~\ref{sec:bootstrapping}, each \plantrace{} and \planimproved{} model is evaluated by generating 64 token sequences for each Sokoban test task.
For the same test task the same model can generate sequences that end in an optimal plan, an incorrect plan, or a correct but sub-optimal plan.
In~\autoref{fig:searchformer-scatter} we compare the length of the generated sequences with the length of sequences generated when using $A^*$ search for each case.
The percentages in each panel's caption list how many out of all generated sequences end in an optimal plan, a correct plan (that can be optimal or sub-optimal), and a correct but sub-optimal plan.

In the left panel of~\autoref{fig:searchformer-scatter-optimal} and~\autoref{fig:searchformer-scatter-correct}, points are centered around the diagonal axis indicating that the \plantrace{} models do approximately match the $A^*$ search algorithm in terms of token sequence lengths.
\autoref{fig:searchformer-scatter-optimal} and~\autoref{fig:searchformer-scatter-correct} further confirm the results presented in Sec.~\ref{sec:bootstrapping}: 
With every improvement step, the points move down and below the diagonal line.
This highlights that the improved \planimproved{} models generate token sequences that are shorter than sequences generated with $A^*$ search.
The \planimproved{} has found a method of searching a task to compute an optimal plan in fewer search steps than $A^*$ search uses.

\autoref{fig:searchformer-scatter-correct-but-suboptimal} illustrates what happens when each model generates a correct but sub-optimal plan.
Here, the \plantrace{} model, that is trained to imitate $A^*$, generates trace sequences that are significantly longer than the sequences generated with $A^*$.
This suggests that the model struggles in computing a correct plan and generates too many search steps, ultimately leading in finding a correct but sub-optimal plan.
Similarly, the \planimproved{} models also generate sequences that are longer than the sequences generated with $A^*$ search.
Despite these inefficiencies, our bootstrapping method is still able to improve the model and bring the average sequence length closer to the length of sequences generated with $A^*$ search (right most panel in~\autoref{fig:searchformer-scatter-correct-but-suboptimal}).
While we would desire the trace length to be low in either case, we found that each model generates a correct but sub-optimal plan with less than 5\% chance.
Consequently,~\autoref{fig:searchformer-scatter-correct} shows that the final \planimproved{} model still generates a correct plan with on average fewer search steps than $A^*$ search.
Statistically, the differences between~\autoref{fig:searchformer-scatter-optimal} and~\autoref{fig:searchformer-scatter-correct} are marginal.

\section{Supplemental figures for experiments}\label{app:experiment-figures}

\begin{figure}[h!]
    \centering
    \includegraphics[scale=1]{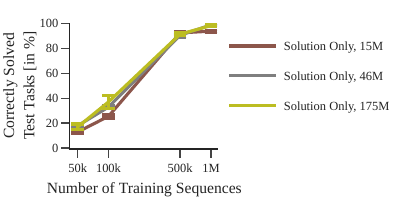}
    \vspace{-3mm}
    \caption{
        \textbf{\Planonly{} model performance.}
        Performance of the \planonly{} models is primarily influenced by the number of training sequences.
        Increasing a model's size does not always improve performance.
    }
    \label{fig:maze-trace-free}
\end{figure}

\begin{figure}[h!]
    \centering
    \subfigure[Deterministic Plan Prediction]{
        \label{fig:trace-free-det}
        \includegraphics[scale=1]{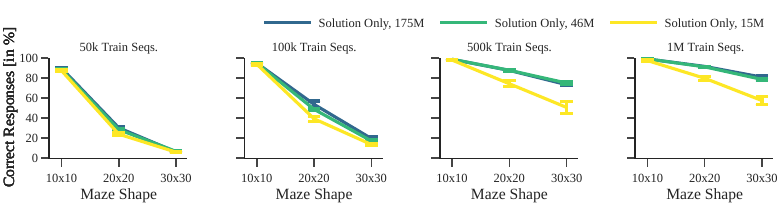}
    }
    \subfigure[Non-deterministic Plan Prediction]{
        \label{fig:trace-free-non-det}
        \includegraphics[scale=1]{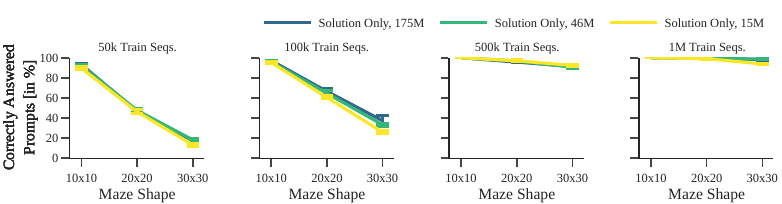}
    }
    \caption{
        \textbf{Optimal plan prediction performance for \planonly{} models.}
        Each panel plots the percentage of correctly answered prompts averaged across five different seeds.
    }
    \label{fig:trace-free-results}
\end{figure}

\begin{figure}[h!]
    \centering
    \subfigure[Deterministic Prediction]{
		\label{fig:trace-added-det}
        \includegraphics[scale=1]{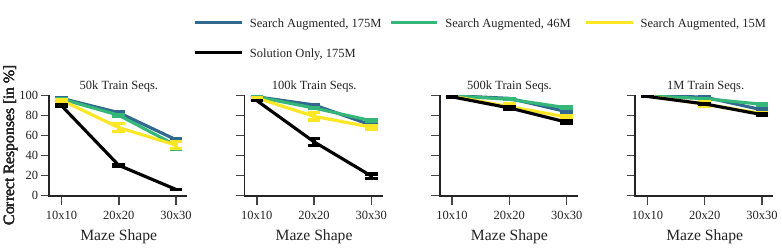}
    }
    \subfigure[Non-deterministic Prediction]{
		\label{fig:trace-added-nondet}
        \includegraphics[scale=1]{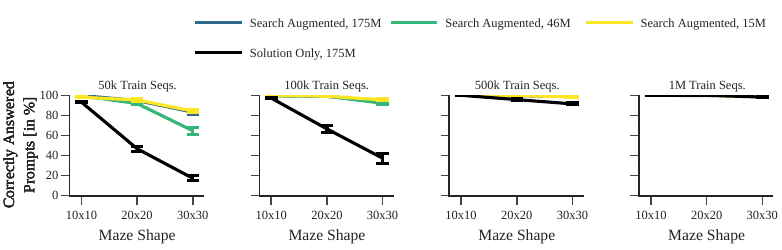}
    }
    \caption{
        \textbf{Optimal plan prediction performance for \plantrace{} models.}
        Training an encoder-decoder Transformer to imitate A* execution traces significantly boosts performance.
        \subref{fig:trace-added-det}: 
        Percentage of correctly generated responses.
        Note that the plan only models only need to generate a few hundred token long plan to answer a prompt correctly.
        The trace plan model must generate a much longer A* execution trace correctly to correctly answer a prompt.
        This requires the trace plan models to generate response sequences that are hundreds or thousands of tokens long (\textit{cf.}~\autoref{fig:sequence-length-stats}).
        If one token is incorrectly predicted, the response is counted as incorrect.
        \subref{fig:trace-added-nondet}: 
        Percentage of correctly generated plans.
        Each model was used to generate multiple responses for each prompt.
        If one of the generated responses contains an optimal plan, the test prompt is counted as correctly answered.
    }
    \label{fig:trace-added-maze}
\end{figure}

\begin{figure}[h!]
    \centering
    \includegraphics[scale=1]{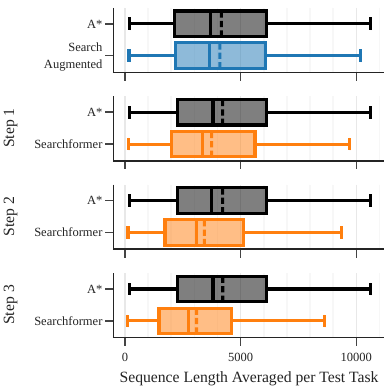}
    \caption{
        Comparison of search dynamics length (in terms of number of tokens) between $A^*$ search and our models (\plantrace{} in blue, \planimproved{} step 1-3 in orange), on the test subset in which our models yield optimal plans.
        Here, for each test task, we average the lengths of sequences that ends in an optimal plan, out of 64 trials (i.e., average-on-optimal length (Appendix~\ref{app:evaluation-criteria})).
        The box plots show the distribution of average-on-optimal lengths, in which the left boundary, mid solid line, right boundary represents the 25\%, 50\% and 75\% quantiles. Dotted lines are means and whiskers show the range.
    }
    \label{fig:searchformer-box-plot}
\end{figure}

\end{document}